\documentclass[10pt,twocolumn,letterpaper]{article}

\usepackage{wacv}              

\newcommand{\cmark}{\ding{51}}%
\newcommand{\xmark}{\ding{55}}%

\definecolor{wacvblue}{rgb}{0.21,0.49,0.74}
\usepackage[pagebackref,breaklinks,colorlinks,allcolors=wacvblue]{hyperref}
\usepackage[accsupp]{axessibility}  

\def\wacvPaperID{230} 
\def\confName{WACV}
\def\confYear{2026}

\title{SCATR: Mitigating New Instance Suppression in LiDAR-based Tracking-by-Attention via Second Chance Assignment and Track Query Dropout}

\author{Brian Cheong \qquad Letian Wang \qquad Sandro Papais \qquad Steven L. Waslander\\
University of Toronto \\
{\tt\small \{brian.cheong, letian.wang, sandro.papais, steven.waslander\}@robotics.utias.utoronto.ca}
}

\begin{document}
\maketitle
\begin{abstract}
    LiDAR-based tracking-by-attention (TBA) frameworks inherently suffer from high false negative errors, leading to a significant performance gap compared to traditional LiDAR-based tracking-by-detection (TBD) methods. This paper introduces SCATR, a novel LiDAR-based TBA model designed to address this fundamental challenge systematically. SCATR leverages recent progress in vision-based tracking and incorporates targeted training strategies specifically adapted for LiDAR. 
    Our work's core innovations are two architecture-agnostic training strategies for TBA methods: Second Chance Assignment and Track Query Dropout.
    Second Chance Assignment is a novel ground truth assignment that concatenates unassigned track queries to the proposal queries before bipartite matching, giving these track queries a second chance to be assigned to a ground truth object and effectively mitigating the conflict between detection and tracking tasks inherent in tracking-by-attention. 
    Track Query Dropout is a training method that diversifies supervised object query configurations to efficiently train the decoder to handle different track query sets, enhancing robustness to missing or newborn tracks. 
    Experiments on the nuScenes tracking benchmark demonstrate that SCATR achieves state-of-the-art performance among LiDAR-based TBA methods, outperforming previous works by 7.6\% AMOTA and successfully bridging the long-standing performance gap between LiDAR-based TBA and TBD methods.
    Ablation studies further validate the effectiveness and generalization of Second Chance Assignment and Track Query Dropout.
    Code can be found at the following link: \href{https://github.com/TRAILab/SCATR}{https://github.com/TRAILab/SCATR}
\end{abstract}    
\section{Introduction}

\begin{figure}
    \centering
    \includegraphics[width=1\linewidth]{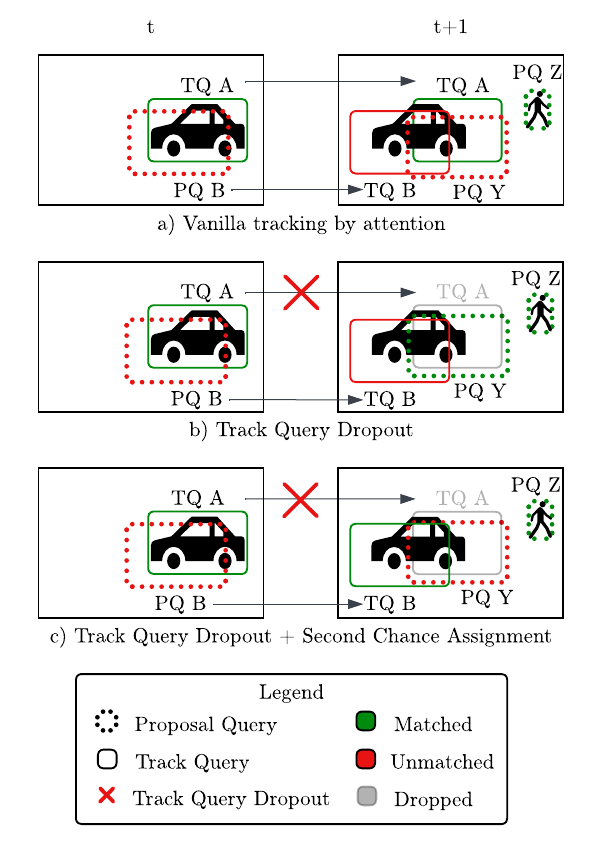}
    \caption{
    a) In regular tracking-by-attention, when a track query is assigned to the vehicle, related proposal queries are suppressed, leading to sparse supervision of proposal queries, only from newborn objects ($PQ\;Z$).
    b) We propose Track Query Dropout, in which assigned track queries may not be propagated to subsequent frames. This provides proposal queries, such as $PQ\; Y$, positive supervision with the now unassigned car, conditioned on the absence of a related track query.
    c) Second Chance Assignment allows both unassigned track queries, like $TQ\;B$, to be assigned to unmatched ground truth objects.
    }
    \label{fig:usecase}
\end{figure}

Multi-object tracking represents a fundamental challenge in computer vision, with critical applications spanning autonomous driving, robotics, and surveillance systems. The task requires accurate object localization across temporal sequences and consistent identity maintenance, making it one of the most complex perception problems in dynamic environments.
The field has traditionally been dominated by tracking-by-detection (TBD) approaches, which decouple the problem into separate detection and association stages. Standard TBD methods utilize established 3D object detectors, such as CenterPoint \cite{yin_center-based_2021} and FocalFormer3D \cite{chen_focalformer3d_2023}, followed by association modules that link sequences of detections into coherent trajectories \cite{yin_center-based_2021, pang_simpletrack_2023}. While TBD methods have achieved remarkable success, particularly in LiDAR-based systems where robust 3D object detectors provide reliable spatial measurements, they inherently suffer from error propagation between stages and limited exploitation of temporal information \cite{zhang_motiontrack_2023, leonardis_jdt3d_2025, gwak_minkowski_2022}.

In contrast, joint detection and tracking (JDT) approaches have emerged as a promising alternative that unifies detection and tracking within a single end-to-end framework \cite{gwak_minkowski_2022, feng_similarity-_2024, yu_motrv3_2023, meinhardt_trackformer_2022, kini_3dmodt_2023}. This paradigm shift is motivated by the proven advantages of end-to-end machine learning approaches in other computer vision domains \cite{mahmoud_dense_2023, hu_planning-oriented_2023, huang_video_2023}. The tracking-by-attention (TBA) paradigm, exemplified by TrackFormer \cite{meinhardt_trackformer_2022} and its successors \cite{yu_motrv3_2023, pang_standing_2023, leonardis_jdt3d_2025, lin_sparse4d_2023}, represents a particular approach to JDT where each track is modelled as a track query with associated information, including feature embeddings and spatial positions. Vision-based TBA methods have demonstrated remarkable success, with approaches such as Sparse4Dv3 \cite{lin_sparse4d_2023} and S2-Track \cite{tang_s2-track_2025} achieving state-of-the-art performance (SOTA) by effectively utilizing temporal context to infer and refine 3D object states over time.

However, a significant performance gap persists between LiDAR-based TBA and TBD methods. Despite LiDAR's inherent precision advantages in 3D measurement over vision methods, existing LiDAR-based TBA approaches, such as JDT3D \cite{leonardis_jdt3d_2025} and MotionTrack \cite{zhang_motiontrack_2023}, have consistently underperformed compared to traditional TBD baselines, including SimpleTrack \cite{pang_simpletrack_2023} and CenterPoint \cite{yin_center-based_2021}. This paradox highlights a fundamental challenge in query-based tracking for LiDAR: new instance suppression.

New instance suppression refers to the under-confidence of queries that detect new objects after the first frame of a sequence. JDT3D \cite{leonardis_jdt3d_2025} showed that while the proposal queries are confident at initializing tracks in the first frame without prior track queries, the proposal query confidence of new objects drops significantly in following frames. The desired behaviour is to have the model identify whether instances in the scene have been previously predicted by a track query and identify the unseen tracks using proposal queries. Instead, the model learns to simply suppress most proposal queries regardless. This critical issue underlying the performance gap between LiDAR-based TBA and TBD arises from the rarity of newborn track cases, leading to the frequent suppression of proposal query detections even if the proposals are good detections, as illustrated in \cref{fig:usecase} a). The conflict between detection and tracking leads to increased false negative errors \cite{leonardis_jdt3d_2025, zhang_motiontrack_2023}.

To systematically address these fundamental limitations and the new instance suppression problem, we introduce SCATR (Second Chance Assignment and Track Query Dropout), a novel LiDAR-based TBA method. SCATR leverages recent advancements from vision-based TBA while incorporating two key innovations in its training strategies specifically adapted for the unique characteristics of LiDAR data, illustrated in \cref{fig:usecase} b) and c). Our paper contributions are summarized below:

\noindent
\textbf{Track Query Dropout}: A tailored training strategy inspired by Group-DETR \cite{chen_group_2023} that creates auxiliary track query groups during training, intentionally exposing the model to diverse query configurations and improving robustness to missing or switched queries. Unlike existing detection query augmentation methods, our approach specifically targets track queries to address the new instance suppression problem in LiDAR-based tracking.

\noindent
\textbf{Second Chance Assignment}: A novel ground truth assignment approach designed to mitigate new instance suppression. This mechanism gives unassigned track queries a second chance to be assigned to untracked objects rather than only relying on proposal queries for new instance detection, improving the detection of newborn instances and ensuring track queries receive supervision for both track continuation and initialization scenarios.

\noindent
\textbf{SCATR model}: We demonstrate the effectiveness of these training strategies using a novel LiDAR-based TBA model that integrates vision-based TBA design with a LiDAR-specific backbone and decoder layers, achieving 65.0\% AMOTA on the nuScenes test split \cite{caesar_nuscenes_2020}

By systematically addressing the new instance suppression problem, SCATR achieves SOTA performance among LiDAR-based JDT methods on the nuScenes tracking benchmark \cite{caesar_nuscenes_2020}. Crucially, our approach successfully bridges the long-standing performance gap between LiDAR-based TBA and TBD, proving the efficacy of these targeted training strategies rather than relying on architectural complexity. This advance paves the way for future end-to-end LiDAR tracking systems that can fully exploit the temporal information available in sequential point cloud data.
\section{Related Works}

\subsection{Tracking-by-Attention}
Multi-object tracking has traditionally been dominated by tracking-by-detection (TBD) approaches, which decouple the problem into distinct detection and association stages. In contrast, tracking-by-attention (TBA) approaches unify detection and tracking within a single end-to-end framework, driven by advances in end-to-end machine learning paradigms. TBA emerged from the foundational work of TrackFormer \cite{meinhardt_trackformer_2022}, which extended the query-based DETR framework \cite{carion_end--end_2020} to multi-object tracking by formulating tracking as a frame-to-frame set prediction problem. Track queries are used to consistently predict distinct tracks in space and time, while proposal queries initialize new tracks \cite{meinhardt_trackformer_2022}. This work established the conceptual foundation for query-based tracking by demonstrating that attention mechanisms could achieve data association without explicit graph optimization or motion modelling.

Building upon this foundation, several vision-based methods extended the tracking-by-attention paradigm to 3D scenarios. PF-Track \cite{pang_standing_2023} proposed an end-to-end multi-camera 3D tracking framework that integrates both past and future reasoning for tracked objects, employing ``Past Reasoning" modules to refine tracks by cross-attending to queries from previous frames and ``Future Reasoning" modules to predict robust trajectories. MUTR3D \cite{zhang_mutr3d_2022} introduced 3D track queries to model coherent tracks across multiple cameras and frames. STAR-Track \cite{doll_str-track_2024} developed adaptive spatio-temporal appearance representations with learnable track embeddings to model existence probabilities, while DQTrack \cite{li2023end} addressed the inherent representation conflict between detection and tracking by decoupling queries into separate object and track components. Sparse4Dv3 \cite{lin_sparse4d_2023} achieves state-of-the-art vision-only tracking performance by extending query denoising to a multi-frame training scheme.
 
The extension to LiDAR-based TBA methods has shown promise but faces significant challenges. 
MotionTrack \cite{zhang_motiontrack_2023} proposed an end-to-end framework that combines transformer-based data association with query enhancement modules, while JDT3D \cite{leonardis_jdt3d_2025} introduced temporally consistent LiDAR track sampling.
While these works represent early attempts to apply transformer-based architectures to LiDAR-based TBA, they failed to demonstrate performance comparable to traditional tracking-by-detection baselines such as SimpleTrack \cite{pang_simpletrack_2023} and CenterPoint \cite{yin_center-based_2021}, suggesting that SOTA vision-based TBA techniques are not directly applicable to the LiDAR domain.
Our work bridges this performance gap by leveraging novel TBA-specific training techniques rather than expanding architectural complexity.


\subsection{New Instance Suppression}
A critical challenge in both vision- and LiDAR-based TBA approaches is the conflict between detection and tracking objectives, leading to the constant suppression of new instance predictions. Prior works explored various strategies to mitigate this suppression by focusing on improving the proposal query detection confidence. In 2D TBA, MOTRv2 used an object detector to generate proposals as anchors to ease the conflict between detection and association tasks \cite{zhang_motrv2_2023}, while MOTRv3 introduced a release-fetch supervision strategy to balance label assignment between detect and track queries \cite{yu_motrv3_2023}.
In vision-based 3D TBA, OneTrack \cite{leonardis_onetrack_2025} attempted to mitigate task conflict by stopping gradients from supervising conflicting queries.
DQTrack \cite{li2023end} took a different approach by using track queries exclusively for tracking while object queries handle detection, thereby eliminating the direct conflict between tasks.
Interestingly, Sparse4Dv3 \cite{lin_sparse4d_2023} achieves TBA without explicit tracking supervision and instead relies solely on detection supervision, assuming that queries will naturally predict consistent objects at inference time. This suggests that for vision-based approaches, new instance suppression might be significantly less of an issue compared to LiDAR-based TBA approaches.
Beyond the fundamental task conflict, JDT3D \cite{leonardis_jdt3d_2025} identified that the desired decoder behaviour is inherently complex and highly dependent on the propagated track queries, making it difficult to determine which query should be responsible for tracking specific objects. This dependency creates a challenging optimization landscape where the decoder must learn context-dependent behaviours.
Our proposed Track Query Dropout efficiently learns these context-dependent behaviours, while Second Chance Assignment uses unassigned track queries to improve the detection of newborn tracks.
 

\subsection{Auxiliary Queries for Detection}
Auxiliary queries have proven to be a highly effective strategy for enhancing the training of detection transformers. Group-DETR \cite{chen_group_2023} introduced multiple independent groups of object queries during training, providing richer supervision and accelerating convergence. Denoising-based methods like DN-DETR \cite{li_dn-detr_2022} and DINO \cite{zhang_dino_2022} improve convergence by injecting perturbed ground-truth queries as auxiliary training targets, effectively stabilizing bipartite matching and improving robustness to assignment instability. Sparse4Dv3 \cite{lin_sparse4d_2023} extends denoising to multi-frame training, providing robust auxiliary training signals over multiple time steps and iterations. While these detection query augmentations improve detection performance and convergence, Track Query Dropout is a query augmentation method that addresses the instance suppression problem unique to TBA, targeting the track query propagation mechanism.
\section{Method}

\begin{figure*}
    \centering
    \includegraphics[width=\textwidth]{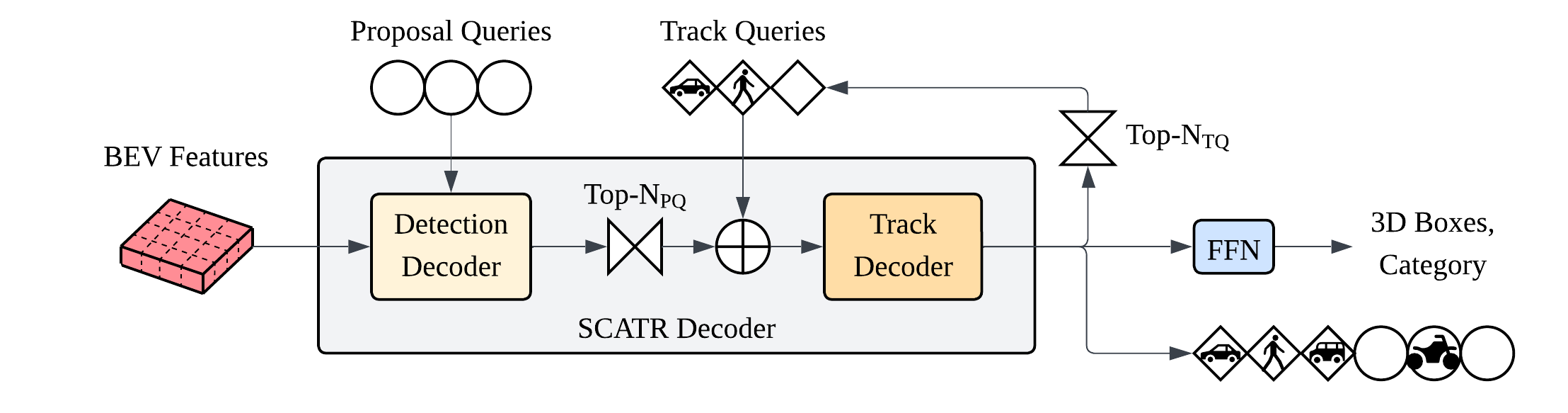}
    \caption{Architecture diagram of SCATR. Given BEV features of the scene and a set of anchor proposal queries, a detection decoder performs an initial detection of objects in the scene. The top-$N_{PQ}$ proposal queries are concatenated with the track queries from the previous frame and passed to the track decoder. The top-$N_{TQ}$ output queries are propagated to the next frame as track queries. Second Chance Assignment allows both track queries and proposal queries to initialize tracks.}
    \label{fig:arch}
\end{figure*}

SCATR adopts the established 3D TBA paradigm \cite{leonardis_jdt3d_2025}, processing sequential LiDAR point clouds using a transformer decoder that jointly performs detection and tracking, using track queries to predict bounding boxes for consistent objects over multiple iterations. 
The core innovations lie in addressing the inherent limitations of LiDAR-based TBA methods through novel, targeted training strategies: Track Query Dropout and Second Chance Assignment. The overall SCATR architecture is summarized in \cref{fig:arch}.

\subsection{LiDAR Backbone \& Query Initialization}

The SCATR pipeline begins with a LiDAR backbone network that encodes raw point clouds into a BEV feature representation. This transformation follows established practices in 3D perception \cite{chen_voxelnext_2023, yan_second_2018, bai_transfusion_2022, chen_focalformer3d_2023, lang_pointpillars_2019}, where the BEV representation provides computational efficiency while preserving spatial relationships critical for tracking. 
Proposal queries are initialized using anchor boxes derived from geometric clustering algorithms, and further refined during training \cite{lin_sparse4d_2023}. 
This anchor-based initialization provides explicit spatial priors that improve convergence compared to randomly initialized learnable queries.

\subsection{Two-Stage Temporal Transformer Decoder}

SCATR employs a two-stage temporal transformer decoder architecture, inspired by Sparse4Dv2 \cite{lin_sparse4d_2023-1} and Sparse4Dv3 \cite{lin_sparse4d_2023}. This staged approach addresses the complexity of joint detection and tracking by decomposing the task into two decoder stages: a detection decoder and a track decoder.
Unlike previous LiDAR-based TBA methods that used shallower decoders \cite{zhang_motiontrack_2023, leonardis_jdt3d_2025}, SCATR employs a deep decoder architecture, providing the model with sufficient capacity to learn complex query handling.

The detection decoder stage focuses exclusively on object detection using proposal queries from the current frame. This stage performs cross-attention between proposal queries and BEV features to generate initial object candidates. By isolating the detection task, this stage can optimize for object localization and classification without the complexity of temporal association.

The track decoder stage integrates temporal information by concatenating the top-$N_{pq}$ most confident proposal queries from the detection decoder with track queries propagated from the previous frame. The combined query set undergoes self-attention to model inter-query relationships, followed by cross-attention with BEV features for spatial grounding. This design enables the model to reason about object persistence and identity across frames. For the first frame of a sequence, when there are no track queries, the top-$N_{pq}$ filtering is skipped, and all proposal queries are passed to the track decoder.



After the two-stage decoding process, the top-$N_{tq}$ track queries with the highest confidence scores are propagated to the next frame. These queries maintain object identities over time by preserving learned representations of tracked objects. This threshold-based propagation naturally creates instances of false-positive track queries and potentially missed tracks. Several propagated track queries may contain temporal information about existing objects, but only one of the queries is assigned to each ground truth object.

\subsection{Track Query Dropout}

\begin{figure}
    \centering
    \includegraphics[width=0.9\linewidth]{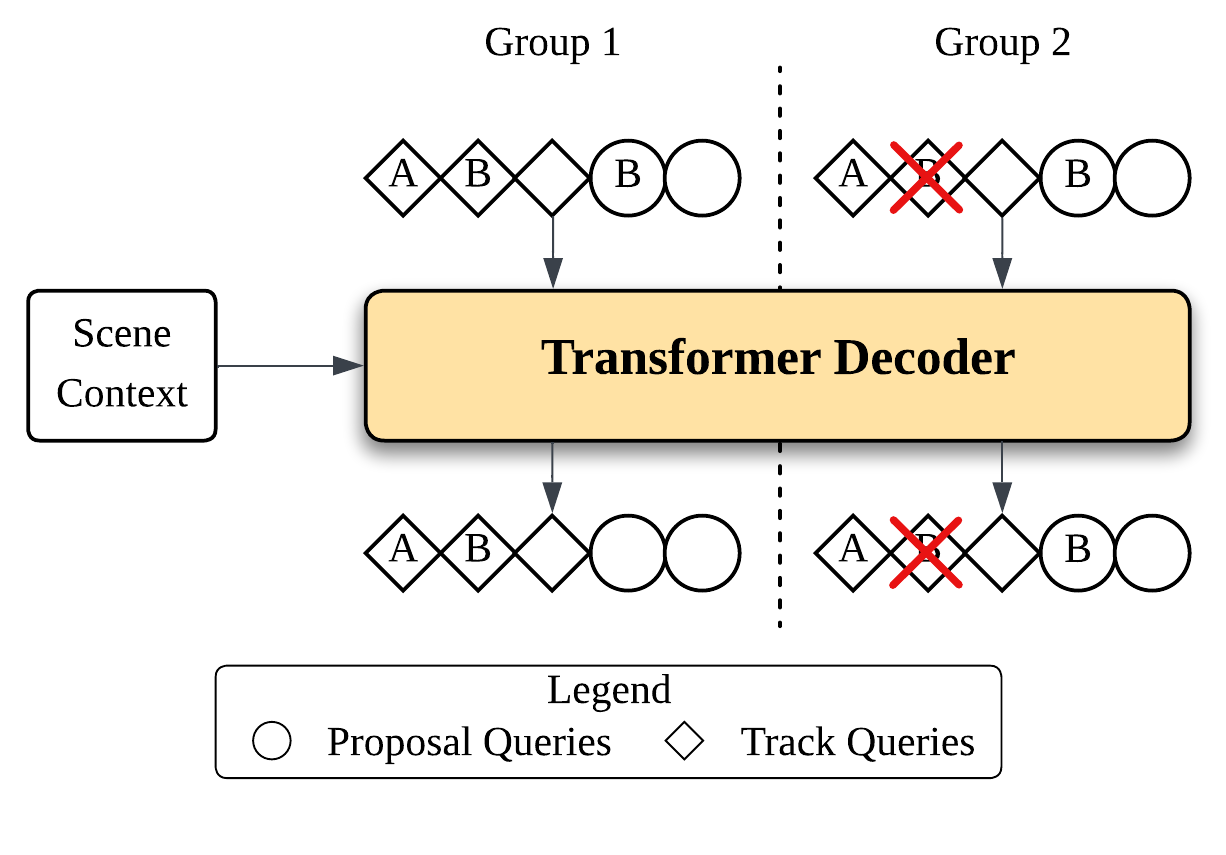}
    \caption{Creating an auxiliary query group by dropping the track query assigned to object $B$ trains the decoder to only suppress the potential proposal query for object $B$ conditioned on whether the assigned track query is present.
    }
    \label{fig:qga}
\end{figure}

SCATR introduces a novel Track Query Dropout strategy to improve training robustness and convergence by efficiently training the model on diverse temporal scenarios.
As illustrated in \cref{fig:qga}, Track Query Dropout prevents some track queries from being propagated to create different ground truth assignments.
The default track query group consists of the top-$N_{tq}$ track queries selected for propagation at each training iteration, following standard TBA \cite{lin_sparse4d_2023, li2023end, tang_s2-track_2025}. Auxiliary groups are formed by randomly sampling $N_{tq}$ queries from the track query set at each training iteration. This operation excludes some assigned track queries while maintaining the same total number of track queries for stacking. The default query group and the auxiliary groups are all propagated to the next sample in the sequence.
By propagating different groups of assigned track queries, the model learns to handle missing or switched track queries.
The behaviour of the decoder in handling proposal queries should be conditional on which track queries were propagated.
For example, \cref{fig:usecase} b) shows an example of dropping $TQ\;A$ from time $t$ to $t+1$, resulting in a modified assignment of the car to $PQ\;Y$. This dropout scenario trains the model to attend to $PQ\;Y$ if $TQ\;A$ is not propagated to the next frame, while the default top-$N_{tq}$ group suppresses $PQ\;Y$ due to $TQ\;A$ being the correct match.
Training with multiple track query groups improves tracking consistency, robustness, and generalization in challenging scenarios such as missed detections under occlusion.

Similar to Group-DETR \cite{chen_group_2023}, all query groups are stacked along the batch dimension before self-attention operations, preventing information leakage between groups while enabling efficient parallel processing. This implementation maintains computational efficiency while providing richer supervision signals. 

\subsection{Second Chance Assignment}

\begin{figure}
    \centering
    \includegraphics[width=0.9\linewidth]{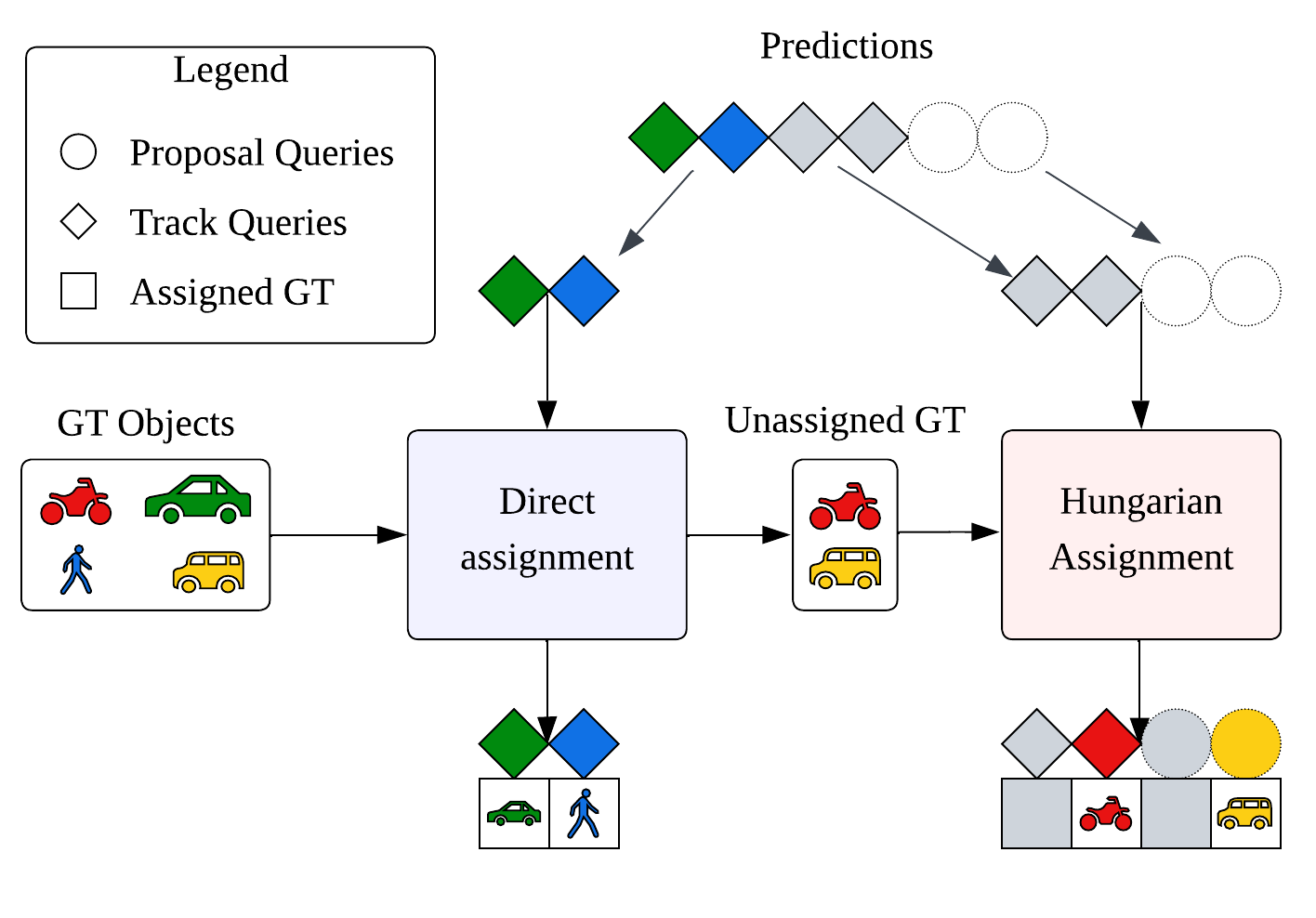}
    \caption{In Second Chance Assignment, unassigned track queries are also considered when matching unassigned ground truth objects, and may take on new track IDs during training. 
    }
    \label{fig:second_chance}
\end{figure}

Previous TBA methods exclusively assign newborn objects to proposal queries, creating an imbalanced supervision regime where proposal queries become under-supervised after initial frames, while track queries become over-confident due to limited exposure to track termination scenarios. 
JDT3D \cite{leonardis_jdt3d_2025} identified this issue as a significant limitation in LiDAR-based TBA systems. 
SCATR addresses this imbalance through a Second Chance Assignment mechanism that balances supervision between detection and tracking tasks, mitigating the inherent conflict within TBA.

Unlike previous TBA methods, unassigned track queries are concatenated to the proposal queries before Hungarian matching with unassigned ground truth objects, as illustrated in \cref{fig:second_chance}. This allows both proposal and track queries to be considered for the assignment of new objects based on their prediction quality and assignment cost, giving the unassigned track queries a ``second chance" to be assigned.
The expanded assignment ensures that track queries receive supervision for both object continuation and object initialization scenarios and leverages the confidence of track queries to detect the rare newborn track cases, reducing false negative rates.

Second Chance Assignment also resolves ambiguities that arise from Track Query Dropout. 
An example is illustrated in \cref{fig:usecase} c), where both $TQ\;A$ and $TQ\;B$ may have information about the vehicle, but only $TQ\;A$ receives the assignment. $PQ\;B$ is propagated and becomes $TQ\;B$, while $TQ\;A$ is dropped. At time $t+1$, the car is now considered to be an "unassigned" in the ground truth assignment since no propagated track queries have been assigned to the car. In traditional TBA assignment schemes, $TQ\;B$ would be suppressed since the assignment would only consider proposal queries, even though $TQ\;B$ may be correctly predicting a high confidence for the car. Second Chance Assignment instead allows $TQ\;B$ to be concatenated to the proposal queries before the bipartite matching and potentially be assigned to the car, since it was not previously assigned a ground truth object.

\subsection{Training and Inference}

During training, the model processes batches of sequential frames, applying the two-stage decoder with Track Query Dropout and Second Chance Assignment. The Hungarian algorithm performs bipartite matching between predictions and ground truth objects, optimizing a combination of classification, localization, and identity consistency losses.

At inference, Track Query Dropout and Second Chance Assignment are not applied.
Only the default track query group is used, maintaining the same computational complexity as standard TBA methods. The model processes frames sequentially, propagating track queries to maintain object identities across time steps. 

\subsection{Integration with Existing Frameworks}

The proposed methods are designed to be generally applicable to various TBA architectures. The two-stage decoder structure can be adapted to different backbone networks and query initialization strategies. Track Query Dropout and Second Chance Assignment can be integrated into existing TBA frameworks with minimal architectural modifications, making them valuable contributions to the broader tracking-by-attention paradigm.

By integrating these components, SCATR addresses fundamental limitations in LiDAR-based tracking-by-attention while maintaining the end-to-end trainability and computational efficiency that make TBA methods attractive for real-world deployment. The method bridges the performance gap between LiDAR-based TBA and TBD approaches by improving training strategies rather than relying on architectural complexity.
\section{Experiments}

\subsection{Dataset}
The nuScenes dataset \cite{caesar_nuscenes_2020} is used for all training and evaluation, following standard splits: 700 training scenes, 150 validation scenes, and 150 test scenes. Each scene contains approximately 40 annotated frames at 2 Hz, with 10 detection and 7 tracking classes.
To evaluate tracking performance, standard tracking metrics from the nuScenes benchmark \cite{caesar_nuscenes_2020} are used, including AMOTA, AMOTP, false positives (FP), false negatives (FN), and ID switches (IDS)\cite{caesar_nuscenes_2020, weng_ab3dmot_2020}. 
The 10-class detection performance is evaluated using mAP.

\subsection{Implementation}
Track Query Dropout is applied before propagating track queries to the next iteration. The main track query group passes the top 600 most confident queries, and the auxiliary group passes 600 random queries as track queries, resulting in a total of two track query groups used during training. 

All experiments were trained using a cyclic learning rate and momentum scheduler with the AdamW optimizer \cite{loshchilov_decoupled_2018}. 
Models were trained using 4 NVIDIA Tesla V100 GPUs for 70,000 iterations and a total batch size of 24, with track sampling and Track Query Dropout disabled after 52,500 iterations. SCATR performs inference at 2.9 Hz on a single V100 GPU.
More implementation details can be found in the supplemental submission.


\subsection{Main Results}
\begin{table*}
\centering
\caption{Tracking performance on the nuScenes \cite{caesar_nuscenes_2020} test split. \textbf{Bold} represents the best performing among Tracking by Attention methods. \underline{Underline} represents the best performing overall. ``-" indicates that a value is not provided by a previous work.}
\begin{tabular}{c|c|ccccccc}
\toprule
\multicolumn{2}{c}{Method}                       & AMOTA $\uparrow$ & AMOTP $\downarrow$ & FP $\downarrow$   & FN $\downarrow$   & IDS $\downarrow$ & mAP$\uparrow$ \\ \hline
Detection & Tracking & \multicolumn{5}{c}{Tracking-by-Detection} \\ \hline
CenterPoint \cite{yin_center-based_2021} & CenterPoint \cite{yin_center-based_2021}        & 0.638 & 0.555 & 18612 & 22928 & 760  & 0.580\\
Centerpoint \cite{yin_center-based_2021}   & SimpleTrack \cite{pang_simpletrack_2023}                & 0.668 & 0.550  & 17514 & 23451 & 575 & 0.580 \\
TransFusion-L \cite{bai_transfusion_2022} & CenterPoint \cite{yin_center-based_2021}  & 0.686 & 0.529 & 17851 & 23437 & 893 & 0.655 \\
VoxelNext \cite{chen_voxelnext_2023} & VoxelNext \cite{chen_voxelnext_2023} & 0.695 & 0.568 & - & - & 785 & 0.645 \\
FocalFormer-L \cite{chen_focalformer3d_2023} & SimpleTrack \cite{pang_simpletrack_2023}              & \underline{0.715} & \underline{0.549} & 16760 & \underline{21142} & 888 & \underline{0.687}  \\
\hline
\multicolumn{2}{c|}{}& \multicolumn{5}{c}{Tracking-by-Attention} \\ \hline
\multicolumn{2}{c|}{PF-Track (camera) \cite{pang_standing_2023}} & 0.434 & 1.252 & 19048 & 42758 & \underline{\textbf{249}} & - \\
\multicolumn{2}{c|}{MotionTrack-L \cite{zhang_motiontrack_2023}}              & 0.51  & 0.99  &   -    &   -    & 9705 & - \\
\multicolumn{2}{c|}{JDT3D \cite{leonardis_jdt3d_2025}} & 0.574 & 0.837 & \underline{\textbf{11152}} & 29919 & 254 & -  \\
\multicolumn{2}{c|}{SCATR (ours)}                        & \textbf{0.650}      & \textbf{0.622}      & 14035      & \textbf{22073}      & 253 & \textbf{0.641}   \\ \bottomrule
\end{tabular}
\label{tab:nusc_test}
\end{table*}

\begin{table*}
\centering
\caption{Tracking performance on the nuScenes \cite{caesar_nuscenes_2020} validation split. \textbf{Bold} represents the best performing among Tracking by Attention methods. \underline{Underline} represents the best performing overall. ``-" indicates that a value is not provided by a previous work.}
\begin{tabular}{c|c|ccccccc}
\toprule
\multicolumn{2}{c}{Method}                        & AMOTA $\uparrow$       & AMOTP $\downarrow$       & FP $\downarrow$   & FN $\downarrow$   & IDS $\downarrow$ & mAP $\uparrow$ \\ \hline
Detection & Tracking & \multicolumn{5}{c}{Tracking-by-Detection} \\ \hline
CenterPoint \cite{yin_center-based_2021} & CenterPoint \cite{yin_center-based_2021}          & 0.637        & 0.606        & 13616 & 22573 & 640 & 0.564  \\
CenterPoint \cite{yin_center-based_2021} & SimpleTrack \cite{pang_simpletrack_2023}                  & 0.687        & 0.573        & 12983 & 19941 & 519  & 0.564 \\
TransFusion-L  \cite{bai_transfusion_2022} & CenterPoint \cite{yin_center-based_2021} & 0.686 & - & 17851 & 23437 & 893 & 0.655 \\
VoxelNext \cite{chen_voxelnext_2023} & VoxelNext \cite{chen_voxelnext_2023} & 0.702 & 0.64 & 12642 & 19179 & 729 & 0.605 \\
FocalFormer-L \cite{chen_focalformer3d_2023} & SimpleTrack \cite{pang_simpletrack_2023}               & \underline{0.728}        & 0.576        & 11039 & 19542 & 528 & \underline{0.664}  \\ \hline
\multicolumn{2}{c|}{} & \multicolumn{5}{c}{Tracking-by-Attention} \\ \hline
\multicolumn{2}{c|}{PF-Track (camera) \cite{pang_standing_2023}} & 0.479 & 1.227 & - & - & 181 & - \\
\multicolumn{2}{c|}{MotionTrack-L \cite{zhang_motiontrack_2023}}                      & 0.617        & \underline{\textbf{0.374}}        &   -    & -      & 20867 & -\\
\multicolumn{2}{c|}{JDT3D \cite{leonardis_jdt3d_2025}}                         & 0.622        & 0.816        &   -    & -      & 203 & 0.399  \\
\multicolumn{2}{c|}{SCATR (ours)}                     & \textbf{0.688}        & 0.609      & \underline{\textbf{10850}} & \underline{\textbf{17556}} & \underline{\textbf{163}} & \textbf{0.645} \\ \bottomrule
\end{tabular}
\label{tab:nusc_val}
\end{table*}

As shown in \cref{tab:nusc_test}, SCATR achieves SOTA performance among LIDAR-based TBA methods on the nuScenes test and validation splits \cite{caesar_nuscenes_2020}. On the test split, SCATR outperforms the previous SOTA TBA method, JDT3D, by 7.6\% on the AMOTA metric, while on the validation split, SCATR outperforms JDT3D by 6.6\% AMOTA and 24.6\% mAP, along with a significant 26\% reduction in false negative errors. 
Our method also compares favourably to baseline TBD methods and outperforms popular methods such as CenterPoint \cite{yin_center-based_2021} and SimpleTrack \cite{pang_simpletrack_2023} on the validation split. PF-Track \cite{pang_standing_2023} is included in \Cref{tab:nusc_test} and \Cref{tab:nusc_val} as a vision-only TBA method for reference.
Several included methods do not have complete publicly available metrics, resulting in the gaps in \Cref{tab:nusc_test} and \Cref{tab:nusc_val}.

Crucially, SCATR successfully narrows the long-standing performance gap between LiDAR-based TBA and TBD methods. SCATR achieves the fewest ID switch errors, the most important metric for track consistency. On the validation split, SCATR has a 19.7\% reduction in IDS relative to past TBA methods and a significant 69.1\% reduction in ID switches compared to TBD methods. This demonstrates the efficacy of our targeted strategies to improve joint training of detection and tracking.

While SCATr significantly narrows the TBA-TBD performance gap, it still falls slightly short on AMOTA and mAP compared to SOTA detectors paired with SimpleTrack \cite{chen_focalformer3d_2023, pang_simpletrack_2023}. This is primarily because these TBD methods leverage highly optimized detection backbones, excelling in raw object localization, especially for minority classes. Despite this, SCATr's value as an end-to-end TBA model is strong: it provides superior tracking robustness with significantly fewer false negatives and ID switches, a crucial benefit for real-world applications. This suggests that while SCATr has substantially advanced TBA, integrating the future LiDAR detection innovations into TBA models may have the potential to further boost performance.

\subsection{Ablations}

\begin{table*} 
\centering
\caption{Ablation on the impacts of second chance assignment and track query dropout on SCATR. ``NB Conf" and ``TQ Conf" refer to the prediction confidence of newborn objects and tracked objects, respectively.}
\begin{tabular}{c|c|cccccccc}
\toprule
S.C. Assignment & TQ Dropout & mAP$\uparrow$      & AMOTA$\uparrow$  & AMOTP$\downarrow$   & FP$\downarrow$   & FN$\downarrow$   & IDS$\downarrow$ & NB Conf$\uparrow$ & Trk Conf$\uparrow$ \\ \hline
\xmark                         & \xmark          & 0.587  & 0.575  & 0.6963  & 11403 & 26490 & 1448 & 0.214 & 0.425 \\
\cmark               &  \xmark         & 0.641  & 0.677 & 0.609 & 11348 & \textbf{17331} & 165 & 0.291 & 0.549 \\
\xmark                         & 1 group & 0.585  & 0.570 & 0.7172  & 12135 & 26403 & 1395 & 0.2143 & 0.420 \\
\cmark               & 1 group & 0.645  & 0.688 & 0.609 & \textbf{10850} & 17556 & \textbf{163} & 0.305 & 0.560\\
\cmark & 2 groups & \textbf{0.648} & \textbf{0.703} & \textbf{0.607} & 11534 & 17579 & 241 & \textbf{0.306} & \textbf{0.565} \\
\bottomrule
\end{tabular}
\label{tab:abl_iter}
\end{table*}


\cref{tab:abl_iter} shows the impact of introducing Second Chance Assignment and Track Query Dropout independently. Applying Second Chance Assignment significantly reduces the number of false negative errors and ID switches, and increases the confidence of newborn and tracked objects without increasing the number of false positive errors. 
By allowing unassigned track queries to be assigned to untracked ground truth tracks, the model no longer needs to rely on the under-confident proposal queries to detect new and missed objects.
As hypothesized, Track Query Dropout alone does not improve tracking performance due to the ambiguities regarding ground truth assignment of the propagated track queries. Applying both described features does slightly improve detection and tracking performance, with further improvements on mAP and AMOTA. Including an extra dropout group yields diminishing improvements at the cost of increased computations during training.

Our targeted training approach substantially improves the overall confidence of true positive newborn object predictions by 9.1\% and tracked object predictions by 13.5\%, as shown by the ``NB Conf" and ``Trk Conf" columns in \cref{tab:abl_iter}, respectively. We also note that previously tracked objects are still being predicted with much higher confidence, similar to previous works \cite{leonardis_jdt3d_2025}. This is expected, as previously seen objects lead to more reliable object tracks.

\subsubsection{Vision Ablations}
\begin{table*}
\centering
\caption{Ablation on the impacts of second chance assignment and track query dropout on SCATR-C using a vision backbone.}
\begin{tabular}{c|c|cccccccc}
\toprule
S.C. Assignment          & TQ Dropout & mAP$\uparrow$      & AMOTA$\uparrow$  & AMOTP$\downarrow$   & FP$\downarrow$   & FN$\downarrow$   & IDS$\downarrow$ & NB Conf$\uparrow$ & TQ Conf$\uparrow$ \\ \hline
\xmark                         &  \xmark         & 0.281 & 0.205 & 1.262 & 13848 & 63567 & 228 & 0.144 & 0.295 \\
\cmark               &  \xmark         & \textbf{0.346} & 0.270 & \textbf{1.156} & 12026 & 52511 & \textbf{73} & \textbf{0.188} & \textbf{0.365} \\
\xmark                         & \cmark & 0.291 & 0.213 & 1.265 & \textbf{9719} & 62262 & 234 & 0.140 & 0.295 \\
\cmark               & 1 group & 0.276 & \textbf{0.288} & 1.165 & 11218 & \textbf{51742} & 92 & 0.186 & 0.363 \\
\bottomrule
\end{tabular}
\label{tab:abl_itercam}
\end{table*}

We also performed ablations to validate whether Second Chance Assignment and Track Query Dropout can also benefit vision-based TBA methods. SCATR was modified for vision input, replacing the LiDAR backbone with a ResNet-50 backbone \cite{he_deep_2016} and using Deformable Feature Aggregation for cross attention \cite{lin_sparse4d_2023-1, lin_sparse4d_2023}. \cref{tab:abl_itercam} shows that applying Second Chance Assignment and Track Query Dropout also significantly improves the detection and tracking performance in the vision domain.



\subsubsection{Number of TQ Dropout Groups}

Experiments were also conducted to explore the impact of the number of dropout groups used during training. 
Our results demonstrate that by concurrently training on a second randomly passed track query dropout group, we achieve improved AMOTA, at the cost of more false positives, false negatives and ID switch errors.
The default configuration of SCATR was set to one track query dropout group since it led to the best performance balance across all metrics with a lower memory footprint.


\subsubsection{TBA vs TBD}
\begin{table*}[!ht]
\centering
\caption{Ablation comparing our SCATR tracking-by-attention against a SCATR-like detector with SimpleTrack \cite{pang_simpletrack_2023} tracking-by-detection. * indicates these values were obtained from applying SimpleTrack to the output detections.}
\begin{tabular}{c|cccccccc}
\toprule
Experiment     & mAP$\uparrow$ & AMOTA$\uparrow$ & AMOTP$\downarrow$ & FP$\downarrow$  & FN$\downarrow$    & IDS$\downarrow$ & TQ Recall\\ \hline
SCATR+det. loss+SimpleTrack & \textbf{0.652} & \textbf{0.702}*     & 0.673* & \textbf{10724}* & 19281*   &  671* & 0.177  \\
SCATR TBA                  & 0.645 & 0.688 &  \textbf{0.609} & 10850 & \textbf{17556} & \textbf{163} & 0.996 \\
\bottomrule
\end{tabular}
\label{tab:abl_tbd}
\end{table*}

\cref{tab:abl_tbd} presents a comparison between SCATR as a TBA method and SCATR trained as an object detector only, using SimpleTrack's approach \cite{pang_simpletrack_2023} to association. The detection loss assumes no track consistency, with all proposal and propagated track queries passed to bipartite matching at each frame, similar to StreamPETR \cite{wang_exploring_2023}.
Despite the increased AMOTA and mAP of the TBD variant of SCATR, the TBA approach still demonstrates tracking robustness with fewer false negatives and ID switches, maintaining consistent IDs through learned temporal reasoning.

Interestingly, without enforcing the assignment of track queries to consistent object instances over multiple iterations, SCATR does not implicitly learn to perform TBA, unlike Sparse4Dv3 \cite{lin_sparse4d_2023}. 
This can be seen in the track query recall, measuring how many tracked objects were assigned to track queries rather than proposal queries.
Training SCATR as a detector results in a track query recall of 17.7\%, showing that a majority of the tracked object assignments are going to the proposal queries rather than the track queries.
One potential explanation is that the LiDAR data input is sparser and more view-dependent than vision data, leading to the reliance on more up-to-date proposal queries rather than the propagated track queries. 
These results show that the supervision of consistent query and track pairs is crucial for LiDAR TBA.

\section{Conclusion}

This work introduced SCATR, a LiDAR-based tracking-by-attention framework that systematically addresses the longstanding new instance suppression problem inherent to query-based tracking. Through Track Query Dropout and Second Chance Assignment, SCATR efficiently disentangles the conflict between proposal and track queries, enriching supervision and enabling the decoder to robustly handle diverse tracking scenarios. Results on the nuScenes benchmark show that SCATR both advances state-of-the-art performance for LiDAR-based TBA and decisively narrows the gap to SOTA tracking-by-detection baselines, yielding marked reductions in false negative and ID switch errors. These results affirm that well-designed training strategies can have a major impact on TBA performance, with benefits in line with recent architectural advancements, paving the way for more unified end-to-end object tracking solutions. 
As TBA continues to mature, promising directions remain in areas such as multi-modal fusion and more adaptive query management for complex and crowded scenes. 
While these are exciting avenues for future exploration, the advances in this work establish a new benchmark for LiDAR-based TBA, setting the stage for broader adoption and continued innovation in the field.



{
    \small
    \bibliographystyle{ieeenat_fullname}
    \bibliography{zotero}
}
\clearpage
\section{Supplemental}
\label{sec:supp}

\subsection{Implementation Detils}
Our implementation is built upon the MMDetection3D library \cite{mmdetection3d_contributors_openmmlabs_2020}.
We leverage the Sparse4Dv3 codebase \cite{lin_sparse4d_2023} to implement query-based tracking, ensuring compatibility with state-of-the-art vision-based tracking-by-attention methods.
For feature extraction, we employ the SECOND LiDAR backbone pretrained on nuScenes for detection provided by FocalFormer3D \cite{chen_focalformer3d_2023, yan_second_2018}. This pretrained backbone is used as a frozen BEV feature extractor during training to provide high-quality spatial representations.

To enhance generalization and robustness, standard LiDAR data augmentations, such as rotations, scaling, and flipping, are applied to the LiDAR sequences in a temporally consistent manner. Track sampling augmentation \cite{leonardis_jdt3d_2025} is also applied to introduce temporally consistent synthetic object tracks to each sequence, increasing the diversity of training scenarios. Temporal instance denoising \cite{lin_sparse4d_2023} is used during training to improve the convergence by adding denoising query groups that are propagated over multiple iterations.

Proposal queries are initialized using 900 anchor boxes derived from k-means clustering and are further refined through training \cite{lin_sparse4d_2023}.
Temporal instance denoising, as described in Sparse4Dv3 \cite{lin_sparse4d_2023}, provides additional auxiliary queries with known ground truth assignments to improve convergence.
Following the detection decoder, the top 300 most confident proposal queries are appended to the track queries before the track decoder.
Track Query Dropout is applied before propagating track queries to the next iteration. The main track query group passes the top 600 most confident queries, and the auxiliary group passes 600 random queries as track queries, resulting in a total of two track query groups used during training. 
Deformable attention is used for all cross-attention operations between the queries and the BEV features \cite{zhu_deformable_2021, chen_focalformer3d_2023}. 
The detection decoder consists of one transformer decoder layer, while the tracking decoder consists of five layers \cite{lin_sparse4d_2023}.

To train SCATR sequentially \cite{wang_exploring_2023, lin_sparse4d_2023}, each training scene is divided into clips of up to 10 annotated frames.
For inference, we use the full sequences of 40 frames.
A custom data sampler handles the batching of sequences of varying lengths.
To improve minority class performance, we use a modified class-balanced data sampling method \cite{zhu_class-balanced_2019, mmdetection3d_contributors_openmmlabs_2020}.
Rather than using the set of classes present in each frame to sample individual frames in a class-balanced manner, the set of classes present in each clip is used to sample subsequences in a class-balanced manner. This ensures that the sequential data loading is maintained rather than having individual frames of a sequence sampled multiple times.

All experiments were trained using a cyclic learning rate and momentum scheduler with the AdamW optimizer \cite{loshchilov_decoupled_2018}. 
Models were trained using 4 NVIDIA Tesla V100 GPUs for 70,000 iterations and a total batch size of 24, with track sampling and Track Query Dropout disabled after 52,500 iterations. SCATR performs inference at 2.9 Hz on a single V100 GPU.

\subsection{Qualitative Results}

BEV visualizations on the nuScenes mini split can be found at \href{{https://drive.google.com/drive/folders/1_l5nF5_qw_I-CAjFHL0BtPf9xI1UphQN?usp=sharing}}{https://drive.google.com/drive/folders/1\_l5nF5\_qw\_I-CAjFHL0BtPf9xI1UphQN?usp=sharing}, showing the detection and tracking performance of SCATR. Track IDs are illustrated with consistent colours in the sequence, and ground truths are illustrated with dashed black boxes.




\end{document}